\documentclass{article}

\usepackage{PRIMEarxiv}

\usepackage[utf8]{inputenc} 
\usepackage[T1]{fontenc}    
\usepackage{hyperref}       
\usepackage{url}            
\usepackage{booktabs}       
\usepackage{amsfonts}       
\usepackage{nicefrac}       
\usepackage{microtype}      
\usepackage{lipsum}
\usepackage{fancyhdr}       
\usepackage{graphicx}       
\graphicspath{{media/}}     

\title{Model-Agnostic Interpretation Framework in Machine Learning: A Comparative Study in NBA Sports
}

\author{
  Shun Liu$^{*}$ \\
  School of Information Management and Engineering \\
  Shanghai University of Finance and Economics \\
  Shanghai\\
  \texttt{kevinliuleo@gmail.com} \\
}

\begin{document}
\maketitle


\begin{abstract}
The field of machine learning has seen tremendous progress in recent years, with deep learning models delivering exceptional performance across a range of tasks. However, these models often come at the cost of interpretability, as they operate as opaque "black boxes" that obscure the rationale behind their decisions. This lack of transparency can limit understanding of the models' underlying principles and impede their deployment in sensitive domains, such as healthcare or finance.
To address this challenge, our research team has proposed an innovative framework designed to reconcile the trade-off between model performance and interpretability. Our approach is centered around modular operations on high-dimensional data, which enable end-to-end processing while preserving interpretability. By fusing diverse interpretability techniques and modularized data processing, our framework sheds light on the decision-making processes of complex models without compromising their performance.
We have extensively tested our framework and validated its superior efficacy in achieving a harmonious balance between computational efficiency and interpretability. Our approach addresses a critical need in contemporary machine learning applications by providing unprecedented insights into the inner workings of complex models, fostering trust, transparency, and accountability in their deployment across diverse domains.
\end{abstract}


\section{Introduction}
\label{sec-intro}
In recent years, the field of sports analytics has experienced a significant transformation with the availability of high-dimensional data. This wealth of data promises to provide profound insights into athlete performance and strategic dynamics in sports. However, one major challenge that persists is the interpretability of complex statistical models used in sports analytics.

Previous studies have made notable contributions to understanding high-dimensional sports statistics. For example, \cite{b19} focuses on identifying players' performance profiles throughout regular and playoff seasons using a combination of tracking and notation variables. By employing clustering algorithms, they categorize players into different clusters based on their performance characteristics. However, this approach may overlook the influence of team quality, as players from powerful teams tend to have higher performance measures.
Additionally, \cite{b18} employs machine learning and deep learning techniques to predict player performance. They use the CRISP-DM methodology and identify the top relevant features in their regression model. On the other hand, \cite{b20} develops a mini-batch optimized neural network to predict the Most Valuable Player (MVP). Their study successfully incorporates effective predictors into the model, achieving high prediction probabilities.

Despite these advancements, there are still significant challenges in interpreting these sophisticated models. Advanced algorithms, such as neural networks and ensemble methods, often lack transparency and interpretability, as discussed in \cite{b1,b2,b3,b4,b5,b10}. The complex nature of these models hinders our ability to understand the underlying decision-making processes.
Moreover, the expansive feature space in high-dimensional datasets adds another layer of complexity. Feature redundancy and collinearity make it difficult to identify the most influential variables, as highlighted by \cite{b6}. Furthermore, the dynamic and context-dependent nature of sports introduces additional challenges. While studies addressing interpretability in image classification and face recognition exist \cite{b7,b8,b9}, there is a scarcity of research specifically focusing on the interpretability of high-dimensional sports statistics, especially in competitive sports analysis.

To address these gaps, our study proposes a novel framework that leverages the intricacies and grounded semantic information of predictive modeling in sports analytics. This framework aims to enhance interpretability by focusing on feature importance, decision mechanisms, and comparative analysis. By shedding light on the black box nature of high-dimensional sports analytics models, we hope to provide valuable insights and enhance the understanding of athlete performance and strategic dynamics in competitive sports. Drawing insights from a diverse array of recent studies including but not limited to the aforementioned, this paper aims to make a substantive contribution to the ongoing discourse on high-dimensional sports statistics. By elucidating the challenges inhibiting model interpretability, we endeavor to pave the way for a more transparent and interpretable future in sports analytics. The main contributions in this paper can be summarized as follows:
\begin{itemize}
    \item Feature analysis in distributed representation space using interpretable machine learning methods. To this end, we propose a set of practical framework in feature engineering coving dimensionality reduction, feature importance measurement, weight visualization, to address the issues of choosing the most important feature or factors to conduct assessment.
    \item We propose an model-agnostic interpretation framework for high-dimensional sports statistics, to complement the claim that model interpretability is hugely important when it comes to empirical analysis and decision-making process, which can provide a solid foundation for post-hoc analysis and domain-specific explanation.
    \item We present a general analysis pipeline in sports realm. In high-dimensional statistics analysis, predictive modeling and model reasoning are highly valuable. To narrow down the gaps between powerful deep learning(DL) models and grounded analysis, we construct a feature-oriented approach for mainstream models, to synergize the efficiency and interpretability, therefore improving its realistic value.
\end{itemize}
The rest of the paper is organized as the following manner:

\section{Related Work}
\label{sec-related}
\textbf{Interpretable Machine Learning.} 
Numerous works \cite{b40,b41,b42,b43,b44,b45,b46,b47,b48,b49,b50} have contributed to the advancement of interpretable machine learning. To be clear, interpretability for machine learning can be partitioned into two parts: model-specific\cite{b51,b52} and model-agnostic \cite{b40,b41,b42,b43,b44}. Model-agnostic interpretability techniques aim to provide insights into the inner workings of machine learning models without making assumptions about the underlying model structure. These methods are designed to be broadly applicable across diverse types of models, offering a generalized approach to understanding their behavior. On the other hand, model-specific interpretability focuses on developing tailored interpretability techniques that leverage the specific characteristics and architecture of individual model types. By capitalizing on the unique properties of each model, these techniques aim to provide targeted and in-depth explanations for model decisions. \cite{b46, b53} offers a comprehensive overview of neural network interpretability frameworks, providing a valuable resource for practitioners seeking to demystify complex model behaviors. \cite{b54,b55,b56,b57,b58} delves into model-specific interpretability techniques . \cite{b49} revolves around the concept of using diverse counterfactual explanations\cite{b50} to explain the decisions made by machine learning classifiers, it emphasizes post-hoc explanation of machine learning through counterfactual approach. The synthesis of model-agnostic and model-specific interpretability techniques represents a pivotal frontier in the pursuit of transparent and comprehensible machine learning models. 

\textbf{Global Model-Agnostic Methods.}
For global model-agnostic frameworks, \textbf{SHAP} (SHapley Additive exPlanations) values are most common, which aims to explain the predictions of machine learning models. SHAP values provide a comprehensive explanation of the importance of each feature globally, as well as locally for individual predictions. The key idea behind SHAP values is to simulate different combinations of feature subsets to compute the influence of each feature on the prediction, thereby assigning a SHAP value to each feature to explain the model's prediction process.

The calculation of SHAP values involves simulating permutations of features and estimating the contribution of each feature to a prediction by taking weighted averages over different subsets of features. The formula for SHAP value \( \phi_i \) for the i-th feature is given by:

\[ \phi_i = \sum_{S \subseteq N\backslash\{i\}}\frac{|S|!(|N|-|S|-1)!}{|N|!}(f_{S\cup\{i\}}(x) - f_S(x)) \]

where:

\( \phi_i \) represents the SHAP value for the i-th feature,
\( N \) is the total number of features,
\( f_S(x) \) denotes the model's prediction when conditioned on the feature subset S,
\( f_{S\cup\{i\}}(x) \) denotes the model's prediction when conditioned on the feature subset S plus the i-th feature,
\( |S| \) represents the number of elements in the set S.

In addition to SHAP values, there are several other commonly used methods for global interpretability, including feature importance, Local Interpretable Model-agnostic Explanations (LIME), and Partial Dependence Plots (PDPs).

Feature importance is a simple and direct method for global interpretability. It assesses the contribution of each feature to the model's predictions by measuring its impact. For tree-based models such as decision trees or random forests, feature importance scores can be calculated to quantify the importance of each feature.

Local interpretable model-agnostic explanations(\textbf{LIME}) is a local interpretability method that aims to explain individual predictions by fitting a simple interpretable model in the vicinity of a given data point. It generates a set of artificial samples and uses a linear model or other simple models to explain the predictions of these samples, providing local explanations for the original model's predictions. These methods offer different perspectives to understand and explain the predictions of machine learning models, helping users gain insights into the model's behavior and the influence of features.

Partial Dependence Plots(\textbf{PDPs}) are a visual global interpretability method that shows the relationship between features and model predictions. They illustrate how the model's predictions change as a single feature varies, revealing non-linear relationships between features and model outputs. Specifically, for a feature \(x_j\), the partial dependence function can be defined as:

\[ PDP_j(x_j) = \frac{1}{N}\sum_{i=1}^{N}f(x_{\sim j}, x_j) \]

where:

\( PDP_j(x_j) \) represents the partial dependence function for feature \(x_j\),
\( N \) is the number of samples,
\( f(x_{\sim j}, x_j) \) represents the model's predicted output for a given value of feature \(x_j\) while keeping other features fixed (\(x_{\sim j}\)).

Permutation feature importance measures the increase in prediction error of the model after we permute the feature. For a model (f) and a feature matrix (X), the permutation importance of feature (j) is calculated as:

\[ importance(j) = \frac{1}{K} \sum_{k=1}^{K} \sum_{i} \left( \hat{y}{i} - \hat{y}^{(k)}{i} \right)^2 \]

Feature interaction explainer(FIE) aims to explain feature interactions by quantifying the relationship between features and the target prediction. It typically involves visualizing the interaction effects between pairs of features using methods such as partial dependence plots or interaction terms in models like decision trees.

\section{Methodology}\label{method}
In this part, some conventional machine learning algorithms are proposed. Of the main objectives of the experiments, it includes winning percentage prediction in the variation of home/away flag, player performance factor analysis, and moreover, exploratory Q\&A module.
Initially, several ML algorithms are involved: linear regression, neural network, support vector machine, and decision tree/random forest, all of above are classic strategy when we are handling with high-dimension data. Formerly, for a data process pipeline, we need to start with 1) EDA: empirical data analysis, “empirical” means to handle the given dataset from the scratch, often including statistics visualization, correlation checking, descriptive analysis(shape, features types, semantic information), and again, some outlier and na values must be excluded or fairly treated, which is called as denoization work. 2) Data Cleaning: data cleaning refers to different aspects: exclude the abnormal points, convert features types, carry out necessary data merging and data augmentation. 3) Basic visualization: this parts include muti-variate distribution visualization and variables’ correlation report.
For a processing pipeline, the design pattern is to construct a complex of data cleaning, model architecture, post processing, evaluation metrics, etc, and this should be the demo, grounded work has much more fabricates and aspectes to consider. As usual, most intensive work will be data cleaning at the very beginning, real-world statistics are melted with extremety, abnormality, exception and even some errors in diversified forms. For example, as [] illustrated, error in data preparation contains various types:

\subsection{Dataset Description}\label{dataset desc}
To deliver a comprehensive picture of players, the dataset used to perform experiments consists of four components: NBA League Dataset, College Basketball Statistics, and Player Social Impact Quantitative Dataset and Seasonal Advanced Statistics.

The first dataset, NBA league dataset, is collected from NBA official website nba.com, with the range of 50 years from 1980s to 2020s.
To be specific, the dataset is merged by four sections: details/team/player/ranking:

'Details' dataset stores the comprehensive technical indicators of races, including wide-ranging statistics of home team and away team:

\begin{itemize}
\item[$\bullet$] Unique ID: game id and team id
\item[$\bullet$] Team-wise Attributes: team abbreviation and team city, 
\item[$\bullet$] Player-wise Attributes: including player profile(name, nickname) and performance measurement(field goal made(FGA), field goal assisted(FGA), field goal three point assisted(FG3A), defensive rebound(DREB), offensive rebound(OREB), block(BLK), steals(STL), turnovers(TO), personal fouls(PF), and plus-minus value).
\end{itemize}

'Team' dataset represents information about all teams in league, with unique teamID for discrimination:

\begin{itemize}
    \item [$\bullet$] Team Profile: team abbreviation, nickname, team city, the capacity of the team-owned arena.
\end{itemize}

'Player' dataset memorizes a rough portfolio of team players, which includes PLAYER\_NAME, TEAM\_ID, (unique)PLAYER\_ID and (game)SEASON. To be clear, the dataset contains the following indicators:

\begin{itemize}
    \item [$\bullet$] Technical Indicators: G(number of games playerd on the season), W(number of winning games on the season), L(number of loosing games on the season), W\_PCT(the winning percentage)
\end{itemize}

'Ranking' dataset primarily reveals the rating in the league based on geometric locations, mainly divided into the East and the West for the most cases.

\begin{figure}
    \includegraphics[width=5cm]{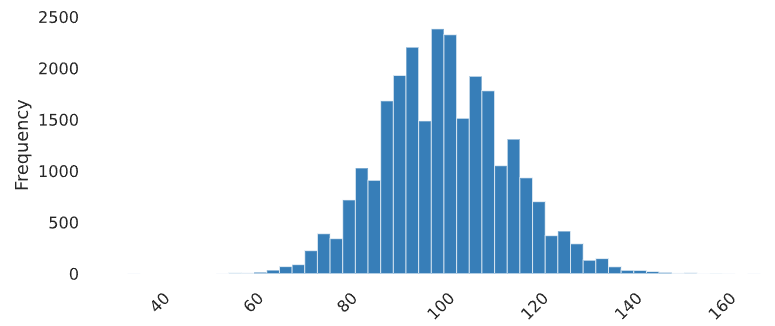}
    \includegraphics[width=5cm]{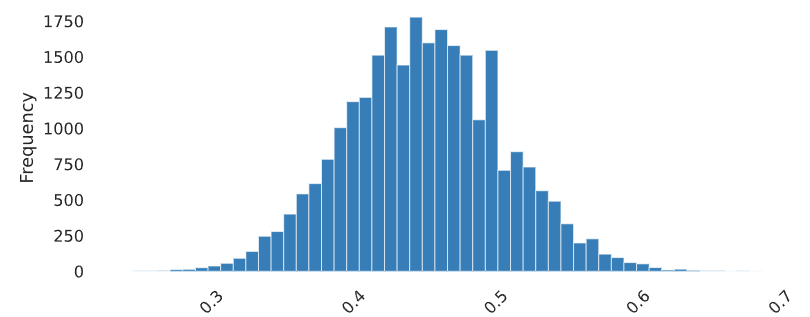}
    \includegraphics[width=5cm]{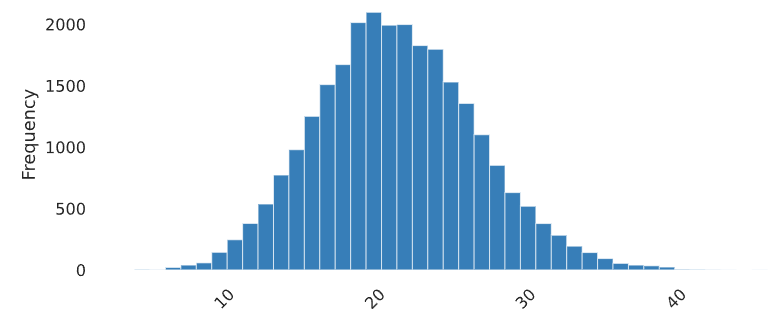}
    \caption{Sample statistics for away team.}
    \label{fig:data-away}
\end{figure}

\begin{figure}
    \centering
    \includegraphics[width=16cm]{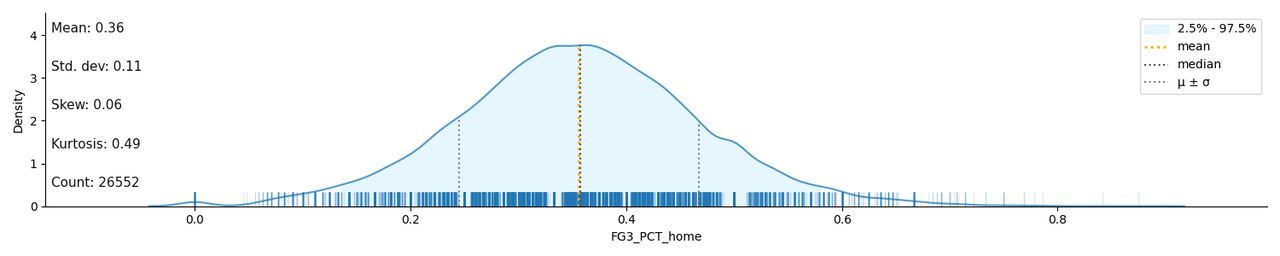}
    \caption{Field Goal Percentage(\%) for home team in the proposed dataset.}
    \label{fig:data-FG3_PCT_home}
\end{figure}

The second dataset used to vaildate "Four-Factors" strategy in Section \ref{lab1} is scraped from https://barttorvik.com, which spans from 2013 to 2023 seasons. The dataset records detailed statistics for teams in college basketball league. Note that the 2020 season is mitigated because COVID'19 epidemic events, hence there is no postseason and the data is sparse, showing the insufficiency and weakness. Note that the size of dataset is approximately 4,000 after the data clearning.

The third dataset targets players' social participation and engagement in a season, which is characterized by mainstream online playform like Twitter, Wikipedia, etc. The quantitative indicators include pageviews, retweats and favorites of posted twitter on average. The pageviews are distinct for each post, so there are 163 thousand unique values for this factor for separate timestamp. Overall, approximately 300 players are recorded in this dataset.

The last dataset is scraped from https://www.basketball-reference.com/, which contains aggregate individual box statistics of multiple seasons, from primary factors like player age, minutes played, in-court role and defensive/offensive box plus/minus, to advanced indicators like win shares/value over replacement/effective field goal percentage. These factors are employed in Section \ref{sec-pos} to make predictions over player role.

\subsection{High-dimensional statistics}



\begin{figure}
    \centering
    \includegraphics[width=15cm]{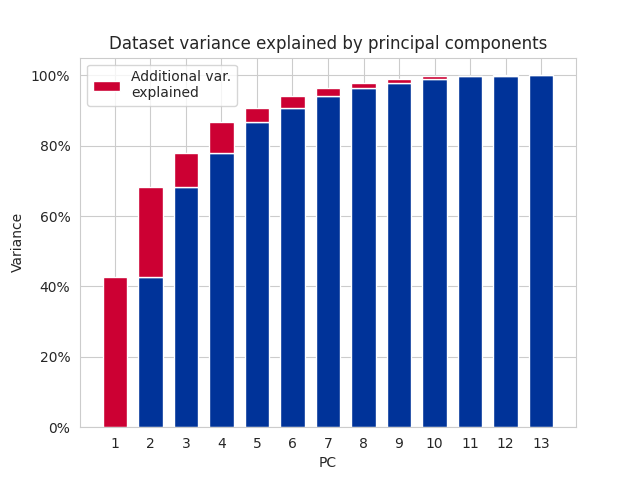}
    \caption{The shifts of dataset variance towards the number of principal components(annotated as n for simplicity). When $X$ equal to 2, with two principal components, the 80\% variance can be explained, in the case of $X$ equals to 3, has better expressiveness(85\% variance is clear). In the subsequent experiments, we prefer $X$ to be 3, which conserve the majority of input's statistical properties, but also reduce information redundency to a larger extent.}
    \label{fig:data-pca}
\end{figure}

There are several algorithms for dimensionality reduction, such as PCA \cite{b33,b38,b39}, MDS \cite{b34}, LDA \cite{b35}, ISOMAP \cite{b36}, and t-SNE \cite{b37}, etc.They focus on filtering out the more representative features among all factors, the "representative" are evaluated on its explanability towards the dataset variances, hence to alleviate the disaster of multicollinearity. For example, with employing principal component analysis(PCA), we could achieve the balance between efficiency and effectiveness. As illustrated in Figure \ref{fig:data-pca}, more principal components can explain more variance, empirically we expect the proposed method to be 85\% explanable after dimension reduction process.

\textbf{PCA:} Principal Component Analysis (PCA) is a statistical method used to simplify the complexity in high-dimensional data while retaining trends and patterns. It achieves this by transforming the original variables into a new set of variables, known as principal components, which are linearly uncorrelated and capture the maximum variance present in the data. PCA is widely employed for dimensionality reduction, visualization of high-dimensional data, feature extraction, and data compression. It is a fundamental tool in exploratory data analysis and has applications across various fields, including finance, signal processing, and image analysis.
    
\textbf{MDS:} Multidimensional scaling (MDS) is a method used to reduce the dimensionality of data and visualize high-dimensional data. Given a data matrix D where the element $d_{ij}$ denotes the distance or similarity between the samples $i$, $j$, MDS tries to find a low-dimensional representation matrix $X$ where the element $x_{ik}$ denotes the coordinates of the sample $i$ in the first $k$ dimension.The goal of MDS is to make the low-dimensional coordinates $x_{ ik}$ preserves the distance relationship between samples as much as possible, minimizing the following loss function:
$$\mathrm{stress}(X) = \sqrt{\frac{\sum_{i \neq j} (d_{ij} - \delta_{ij})^2}{\sum_{i \neq j} d_{ij}^2}}$$
where $\delta_{ij}$ denotes the Euclidean distance between the low-dimensional coordinates $x_{ik}$. By minimizing the loss function, the MDS algorithm can compute the coordinates of the samples in the low-dimensional space based on the distance or similarity matrix, thus realizing data reduction and visualization.

The steps of the algorithm of MDS are as follows:
\begin{enumerate}
    \item Calculate the distance or similarity matrix between the samples $D$: first determine the distance or similarity between the samples, usually use the Euclidean distance, the Ma distance or the correlation coefficient to calculate the distance or similarity between the samples, and construct a distance matrix or similarity matrix $D$ of $n\times n$.
    \item Calculate the Gram matrix $B$ of the samples: Convert the distance matrix $D$ into a Gram matrix $B$ where the element $b_{ij} = -\frac{1}{2}d_{ij}^2$. The Gram matrix can be viewed as an inner product relationship between the samples.
    \item Bilateral Centering: Bilateral centering of the Gram matrix $B$ $Y$ ields the matrix $B'$. This involves performing the following two centering operations on the matrix $B$: first, summing the rows and columns of the matrix $B$ to zero by subtracting the mean of each row and the mean of each column; and second, dividing by the number of samples $n$ to ensure that the trace of the centered matrix is zero.
    \item Eigenvalue decomposition: Perform eigenvalue decomposition on the centered matrix $B'$ to obtain the eigenvector matrix $V$ and the diagonal matrix $\Lambda$, where the column vectors of $V$ are the eigenvectors and the diagonal elements of $\Lambda$ are the eigenvalues.
    \item Select the main eigenvectors: Based on the magnitude of the eigenvalues, the first $k$ eigenvectors are selected to form the matrix $V_k$. These eigenvectors correspond to the largest eigenvalues and represent the main structure of the data set.
    \item Calculate the low-dimensional coordinate matrix $X$: The low-dimensional coordinate matrix $X$ of the sample is calculated as $X = V_k \sqrt{\Lambda_k}$, where $\sqrt{\Lambda_k}$ is the diagonal matrix formed by taking the square root of the first $k$ eigenvalues. These coordinates represent the position of the sample in the low-dimensional space.
\end{enumerate}

\textbf{LDA:} The LDA method calculates the dispersion $S_w$ for the "within-class" and $S_b$ for the "between-class" respectively, and we want $S_b$/$S_w$ to be as large as possible to find the most suitable mapping vector $w$. The LDA method also calculates the dispersion $S_w$ for the "within-class" and $S_b$ for the "between-class" respectively.

\textbf{ISOMAP:} Isomap algorithm is an algorithm derived from MDS algorithm, MDS algorithm is to keep the distance between the samples unchanged after dimensionality reduction, Isomap algorithm introduces the neighborhood graph, the samples are only connected with their neighboring samples, and the distance between them can be calculated directly, and the distance of the more distant points can be calculated by the minimum path, based on which dimensionality reduction to preserve the distance is carried out.

This method works as the following:
\begin{enumerate}
    \item Set the number of points in the neighborhood, calculate the adjacency distance matrix, and set the distance not outside the neighborhood to infinity;
    \item Find the minimum path between each pair of points and convert the adjacency matrix matrix to minimum path matrix;
    \item Input the MDS algorithm to produce the result, which is the result of the Isomap algorithm.
\end{enumerate}

\textbf{t-SNE:} t-SNE utilizes a heavier long-tailed t-distribution in low dimensions to avoid crowding and optimization problems. t-SNE is a nonlinear dimensionality reduction algorithm that is very suitable for visualizing high-dimensional data down to 2 or 3 dimensions. t-SNE was proposed by Laurens van der Maaten and Geoffrey Hinton in 2008. t-SNE is very similar to SNE. The discrepancy are as follows:
\begin{itemize}
    \item t-SNE uses symmetric version of SNE to simplify the gradient formula. One way to optimize the KL dispersion of $p_{i|j}$ and $q_{i|j}$ is to replace the conditional probability distribution with the joint probability distribution, i.e., P is the joint probability distribution of the points in the high-dimensional space, and Q is the low-dimensional space, and the objective function is: $p_{i|j}$ and $q_{i|j}$.
$$C=KL(P||Q)=\sum_i\sum_jp_{i,j}\log\frac{p_{ij}}{q_{ij}},\ p_{ii}=q_{ii}=0$$
    \item In low-dimensional space, the t-distribution is used instead of the Gaussian distribution to express the similarity between two points.
\end{itemize}

\textbf{One-Way ANOVA}, or one-way analysis of variance, is a statistical method used to compare the means of three or more independent (unrelated) groups to determine if there are statistically significant differences between them. This method is often used in research and data analysis to assess whether the means of different groups are equal or not.

The mathematical formula for one-way ANOVA involves several components:

1. Total sum of squares (SST): The total variability in the data, calculated as the sum of the squared differences between each individual data point and the overall mean of all data points.

\[ SST = \sum_{i=1}^{n}\sum_{j=1}^{m}(X_{ij} - \bar{X})^2 \]

where \( X_{ij} \) represents the individual data points, \( \bar{X} \) is the overall mean, \( n \) is the total number of groups, and \( m \) is the number of observations in each group.

2. Between-group sum of squares (SSB): The variability between the group means and the overall mean, calculated as the sum of the squared differences between each group mean and the overall mean, weighted by the number of observations in each group.

\[ SSB = \sum_{i=1}^{n} n_i (\bar{X}_i - \bar{X})^2 \]

where \( n_i \) is the number of observations in the \( i \)th group, \( \bar{X}_i \) is the mean of the \( i \)th group, and \( \bar{X} \) is the overall mean.

3. Within-group sum of squares (SSW): The variability within each group, calculated as the sum of the squared differences between each individual data point and its respective group mean.

\[ SSW = \sum_{i=1}^{n}\sum_{j=1}^{m}(X_{ij} - \bar{X}_i)^2 \]

where \( X_{ij} \) represents the individual data points, \( \bar{X}_i \) is the mean of the \( i \)th group, \( n \) is the total number of groups, and \( m \) is the number of observations in each group.

The F-statistic for one-way ANOVA is then calculated as the ratio of the between-group variability to the within-group variability, adjusted for the degrees of freedom:

\[ F = \frac{SSB / (n-1)}{SSW / (N - n)} \]

where \( n \) is the number of groups, \( N \) is the total number of observations, \( SSB \) and \( SSW \) are the between-group and within-group sum of squares, and \( (n-1) \) and \( (N - n) \) are the degrees of freedom for the between-group and within-group variability, respectively.

The F-statistic follows an F-distribution, and by comparing the calculated F-value to the critical value from the F-distribution for a given significance level, we can determine whether there are significant differences between the group means. If the p-value associated with the F-statistic is less than the chosen significance level (often 0.05), we reject the null hypothesis and conclude that at least one pair of group means are significantly different.

\textbf{Feature Engineering(FE)} is commonly used to select feature according to past experiences and domain knowledge. Usually, a successful feature engineering can balance model performance and dataset explanability, as shown in Figure \ref{dataset desc}. Outstanding feature engineering refers to specialised expertised capability, specialized insights are mostly important. 

\subsection{Preprocessing}
\textbf{Variable Scaling.} In order to regularize the input to a certain range, we have normalized the data in the range from 0 to 1 using the following formula:
$$x_n = \frac{x_i-x_{min}}{x_{max}-x_{min}}$$
where,
$x_n$is the normalized data given the input variable
$x_{min}$indicates the minimum among all input variables
$x_{min}$shows the maximum of the input vairables

\textbf{Multiple Imputation.}

\section{Experienment} \label{analysis}
This section stress the importance of model selection, which paves the way for model interpretability in Section \ref{sec-inter}. In empirical analysis, with a decent amount of candidate machine learning(ML) models, it's uneasy to choose most effective but cost-benefit model in the first glance. To achieve this goal, we aims to conduct fine-grained researches and dive into the model fundamentals so as to explore the grounded performances of the following models: linear regression, neural network, decision tree, and support vector machine.

\subsection{Expecting Player Role Based On Their Advanced Box Statistics}\label{sec-pos}
Player role, also indicates the position of players in a court. For typical line-ups, it refers to five player roles, with its unique functions and responsibilities.

\begin{itemize}
    \item Point Guard:
The point guard is typically the team's primary ball-handler and playmaker, responsible for organizing the offense and distributing the ball. They need to possess excellent passing and ball-handling skills, as well as a keen understanding of the game's flow.
Point guards are often considered leaders on the court, taking on the responsibility of leading the team.
    \item Shooting Guard:
The shooting guard is usually the team's scoring threat, excelling in shooting and scoring.
They require exceptional shooting abilities, quick drives to the basket, and a solid defensive presence.
    \item Small Forward:
Small forwards are versatile players who contribute to scoring and defense.
They are typically of medium height, possessing good shooting, jumping, and athletic abilities, while also demonstrating defensive prowess.
    \item Power Forward:
Power forwards are often the primary inside players, controlling rebounds and the interior on both offense and defense.
They need strong inside scoring, rebounding, and defensive abilities, protecting the rim on the defensive end.
    \item Center:
Centers are the team's interior anchors, responsible for rebounding, scoring, and defending the paint.
\end{itemize}

Typically the tallest players on the team, they need exceptional rebounding and shot-blocking abilities, while also contributing to scoring on offense. While these roles have specific specialties, players often transition between different roles based on the game's requirements. A successful team usually features players in various roles working together to strive for the team's victory.

To derive the correlation between players' height/weight, we have conducted the extensive experiments with interpretable models and manual observations.

\subsubsection{Experiment protocols}
\begin{figure}
    \centering
    \includegraphics[width=8cm]{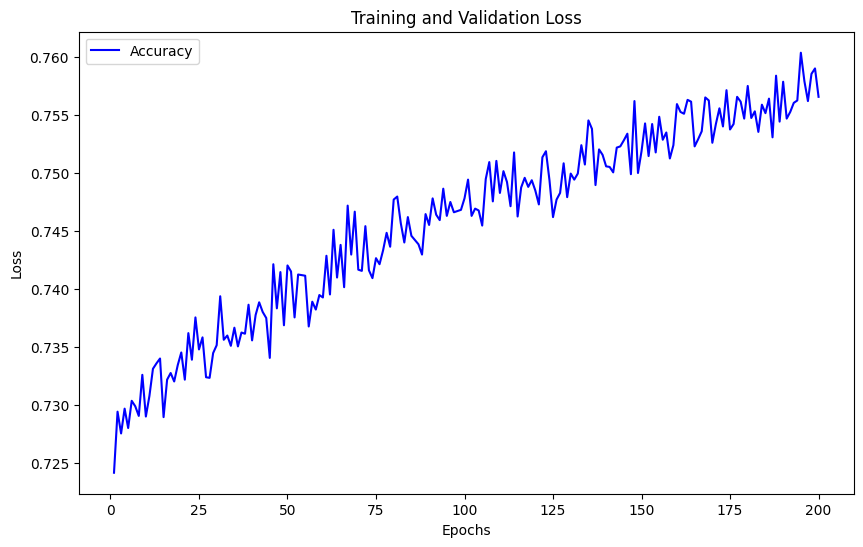}\label{nn-perform}
    \includegraphics[width=8cm]{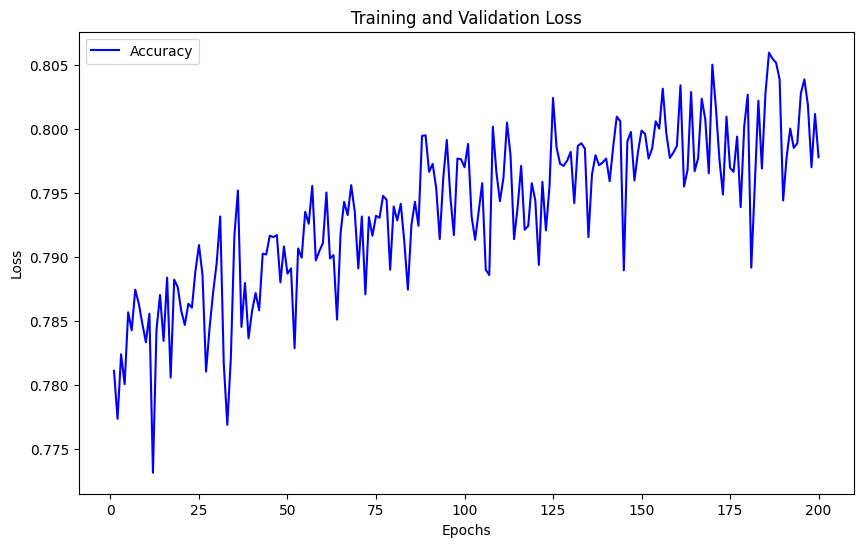}
    \includegraphics[width=12cm]{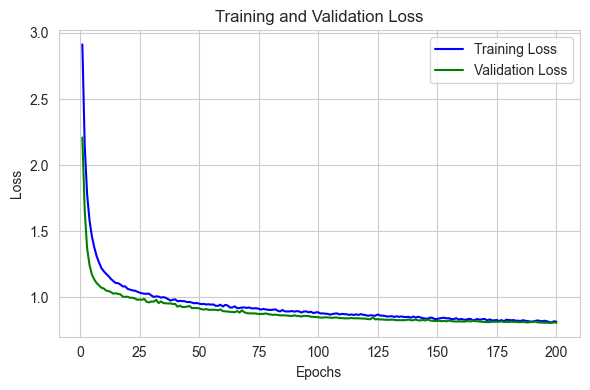}\label{nn-loss}
    \caption{Training losses(left) and retraining accuracy(right) of artificial neural network(ANN). The losses for both training and validating keep declining before plateauing at the range of 0.6-0.8, and the gaps are narrowed with training epochs increasing, indicating the network has converged. Aim to improve the performance, 200 epochs of retrainig is conducted and the scoring metric--accuracy achieves 0.76.}
    \vspace{-10pt}
    \label{nn-loss}
\end{figure}

\textbf{Model.} We propose artificial neural network(ANN) to make predictions. Specifically, 15,378 training and 3,845 validation samples are fed into the model, this neural network consists of three fully connected layers. The first layer has 40 neurons with the $ReLU$ activation function and takes an input of dimension 46. The second layer is a dropout layer with a dropout rate of 0.5, which helps prevent over-fitting by randomly setting a fraction of input units to 0 at each update during training. The third layer has 30 neurons with the $ReLU$ activation function, followed by another dropout layer with a dropout rate of 0.5. Finally, the output layer consists of a Dense layer with a softmax activation function, which is suitable for multi-class classification tasks as it outputs a probability distribution over the different classes. Re-training accuracy and validation performance are visualizaed in Figure \ref{nn-perform}

Of equal importance, hyperparameter tuning, such as learning rate, batch size, and regularization techniques(Lasso/Ridge Regularization), can further refine the model's performance. Furthermore, techniques like early stopping and batch normalization can be applied to improve training efficiency and generalization.

\textbf{Metrics.} Scoring metrics is accuracy, a ratio of predicted positive instances to actual positive instances. With the configured network architecture, validating accuracy maintains stable at the level of 0.77 as shown in Figure \ref{nn-loss}.

Considering that in the NBA league, some players may possess different positions in the career, such as Michael Jordan and Magic Johnson. Recognising the converting patterns of player role required in-depth knowledge and good generalization for the model. In order to prove the robustness of proposed method, this case is involved in the experiment as well. Through further experiment, the model is able to detect the conversion of Jordan into a forward at the end of his career, but not the return of Magic as a power forward. Also, in his rookie season, he is classified as a small forward instead of as a shooting guard (Magic was clearly and outlier in the data, a 205cm point guard who could easily play in the five position. It is even surprised that is properly labelled as a point guard during most of his career). We could make a summary that retraining is effective to boost performance in machine learning realm. To this end, we re-train with other competition seasons excluded in the former experiment, and visualize the results.

\subsection{Predicting The Amount of Winning Games}\label{lab1}
Paradgimic strategy to analyze basketball team performance and efficiency is through box statistics. Standard methods like "Four-Factors" statistics characterize the profiles with four principal indicators: shootings, turnovers, rebounding, and free throws. The semantics information of these factors are displayed in Table \ref{four factor table}. By leveraging these factors, athletes and coaches can gain a comprehensive understanding of a team's performance and identify areas for improvement.
Briefly speaking, the task for this experiment is to validate the effectiveness of 'Four-Factors' evaluation strategy with grounded dataset, and dive into different machine learning models to carry out comparative studies.

\begin{figure}
    \centering
    \includegraphics[width=16cm]{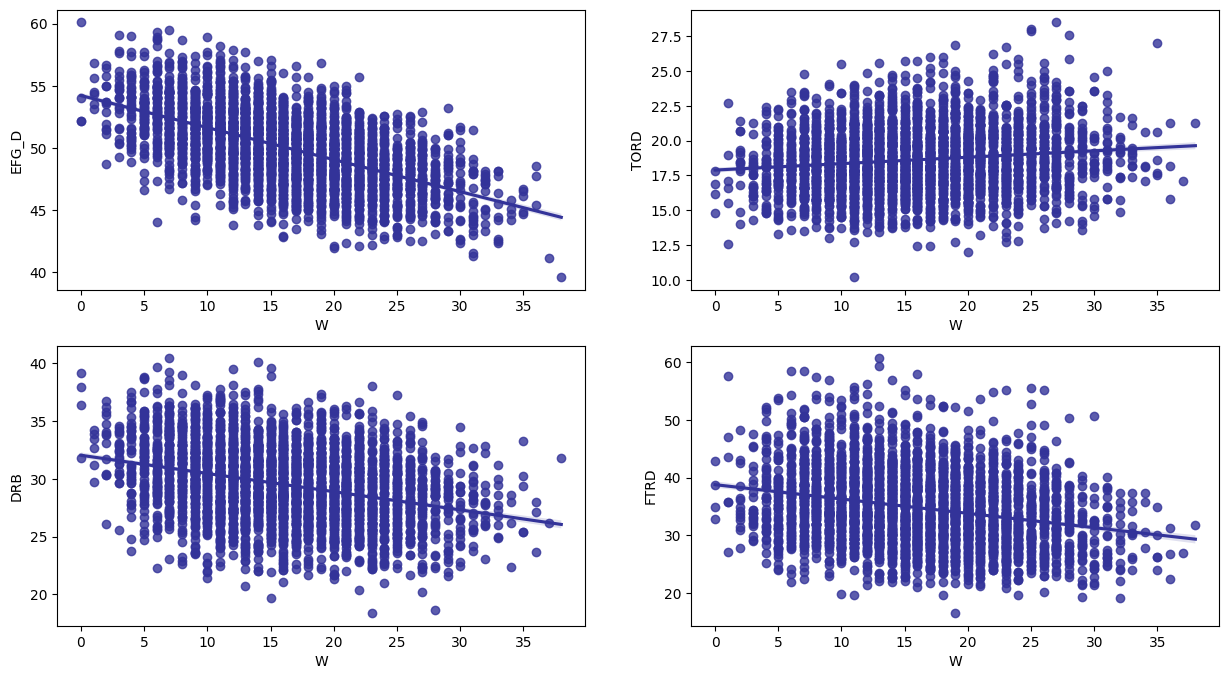}
    \caption{Pair-wise correlation between "Four Factors" and "W"(the number of winning games).}
    \label{four factor}
\end{figure}

As normal, Technical indicators are subject to normal distribution; Regression plot with naive linear regression have been shown in Figure \ref{four factor} to test the pair-wise linearity. The fact is clear that defensive indicators including X and X contribute more linearity to the target variable compared to the offensive facets.

\begin{table}
 \caption{The semantic information of the "Four Factors", which consists of shooting, turnovers, rebounding, and free throws. The "Four Factors" are essential metrics in evaluating a team's performance in the NBA. Shooting and free throws are critical components of a team's offensive output, while turnovers and rebounding are important for both offense and defense. A team's success in these areas can often determine the outcome of the game.}
  \centering
  \begin{tabular}{lll}
    \toprule
    \multicolumn{2}{c}{\textbf{Fundamentals}}                   \\
    \cmidrule(r){1-2}
    \textbf{Factor}     & \textbf{Description}     & \textbf{Grounded Importance} \\
    \midrule
    Shooting & Percentage of shots made  & Essential for scoring points, which is the primary objective0     \\
    Turnovers     &  Possession lost without shot attempt & Critical; can lead to easy baskets for the opposing team      \\
    Rebounding     & Ability to retrieve missed shots       & Preventing easy opponent points  \\
    Free Throws     & Number and accuracy of free throws attempted       & Crucial for scoring points and drawing fouls to gain an advantage  \\
    \bottomrule
  \end{tabular}
  \label{four factor table}
\end{table}

In summary, the "Four Factors" are key metrics in evaluating a team's performance in NBA. Shooting and free throws are critical for scoring, while turnovers and rebounding are important for preventing the opposing team from scoring.

\subsubsection{Candidate Regression Models}
To be specific, we set the 'W' as the target representing the number of winning games as the target variable, and the rest as the independent variables, then frame it as a regression problem. Intuitively, linear regression approaches are employed, here naive linear regression, generalized linear models(GLMs) and their variants(\textbf{\textit{TweedirRegressor}}, \textbf{\textit{HuberRegressor}}) are involved to assure better performance. Validating metrics for the regression problem is $R^2$ score as usual.

\textbf{Linear Models.}
The linear models used for regression tasks are varied in loss function and learning objective. \textbf{\textit{GLMs}}, \textbf{\textit{TweedieRegressor}} and \textbf{\textit{HuberRegressor}} are employed in this experiment.
 
1. \textbf{Ordinary Least Square (OLS).} Ordinary least square is a statistical method commonly used in regression analysis to estimate the unknown parameters of a linear model. It works by minimizing the sum of the squared differences between the observed values and the predicted values from the linear model. OLS assumes that the errors are normally distributed and have constant variance, and it provides estimates of the coefficients of the linear equation that best fit the data. OLS is widely used in many fields, such as economics, finance, and engineering, as it provides a simple and efficient method for estimating linear models and analyzing relationships between variables.

2. \textbf{Generalized Linear Models (GLMs)}, are a class of regression models that extends the traditional linear regression model by allowing for non-normal error distributions and non-linear relationships between the independent and dependent variables. In a GLM, the response variable $Y$ is assumed to follow a distribution from the exponential family, which includes common distributions such as normal, binomial, Poisson, and gamma. The distribution of $Y$ is characterized by its mean $\mu$ and a variance function $V(\mu)$, which relates the mean to the variance. The relationship between the independent variables $X$ and the mean $\mu$ of the response variable is modeled through a link function $g$, such that $g(\mu) = \eta = X\beta$, where $\eta$ is the linear predictor and $\beta$ is a vector of regression coefficients. The link function $g$ maps the mean to the linear predictor, allowing for non-linear relationships between $X$ and $Y$ .

3. \textbf{TweedieRegressor} is a GLM-based algorithm for modeling continuous response variables with a Tweedie distribution. The Tweedie distribution is a two-parameter family of distributions that includes the normal, Poisson, and gamma distributions as special cases. The Tweedie distribution is characterized by a power parameter $p$ and a dispersion parameter $\phi$, which control the shape and variability of the distribution.




4. \textbf{HuberRegressor} is a robust linear regression algorithm that is less sensitive to outliers compared to traditional linear regression models, and does not assume any specific error distribution. Outliers can have a significant impact on the estimated coefficients in a linear regression model, leading to biased and unreliable results. HuberRegressor addresses this issue by minimizing a modified loss function that is less sensitive to outliers.




\subsubsection{Regression Results}
The quantitative results is shown in Table \ref{lab1-table}. \textbf{\textit{NeuralNetwork}} achieves the best performance among all candidate models, with $R^2 score$ 0.8156, followed closely by \textbf{\textit{OLS}}(0.8102) and \textbf{\textit{HuberRegressor}}(0.8046). During the training process, the network is able to converge and get the validation loss around 3.5032.

\begin{table}
 \caption{Performances of the candidate models, within the $R^2 score$  metrics.}
  \centering
  \begin{tabular}{ll}
    \toprule
    Model     & $R^2$ Score(Test Set)      \\
    \midrule
    \textbf{\textit{OLS}} & 0.8102     \\
    \textbf{\textit{Lasso(alpha=0.05)}}     &  0.5437       \\
    \textbf{\textit{Ridge(alpha=0.5)}}     & 0.7861       \\
    \textbf{\textit{TweedieRegressor}}     & 0.7999       \\
    \textbf{\textit{HuberRegressor}}     & 0.8046        \\
    \textbf{\textit{Decision Tree}}     & 0.6049       \\
    \textbf{\textit{Neural Network}}     & \textbf{0.8156}       \\
    \bottomrule
  \end{tabular}
  \label{lab1-table}
\end{table}

\subsection{Bridging Player Performances With Salary}
To dive into the correlation between player performance versus salary, we have conducted several grounded experiment to testify pair correlation with visualization and statistical analysis. To be specific, the involved factors are all-covered, which not only including defensive and offensive quality towards single player, but also deliver a comprehensive picture on player’s past performance and the trending across several league seasons. The testing factors are enlisted in below: \textit{PTS(for points)/DRPM/ORPM}

For out-court indicators, we consider player’s social participation and interpersonal interactions with online world, try to excavate practical relational patterns. Strategically, this part address the knowledge in statistics and machine learning, such as p-testing, linear regression report, etc, to fully understand how social power affects player’s salaries. 

During the experiment, we employ \textit{'PV(', 'TFC(for twitter favorite counting)', TRC(for twitter retweet counting)', 'MPG', 'PTS(for points)', 'DRPM', 'ORPM', 'PN(for position number)', 'AGE'} as given variables and assume \textit{SM(counting the salary in millions))} as the target variable, their relations have been displayed in the Figure \ref{weight plot 2}.

\begin{figure}
    \centering
    \includegraphics[width=8cm]{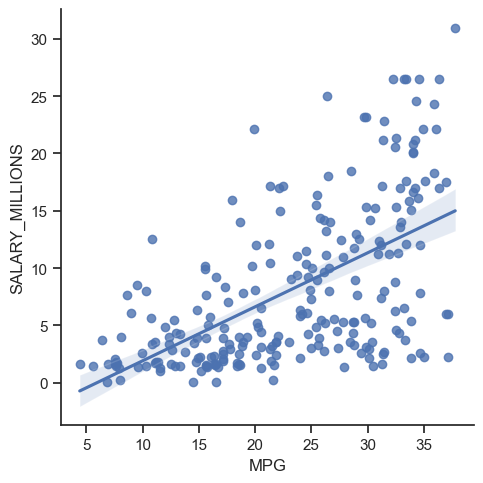}
    \includegraphics[width=8cm]{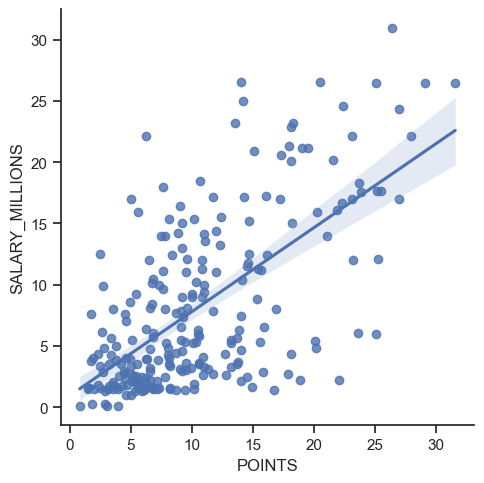}
    
    \includegraphics[width=8cm]{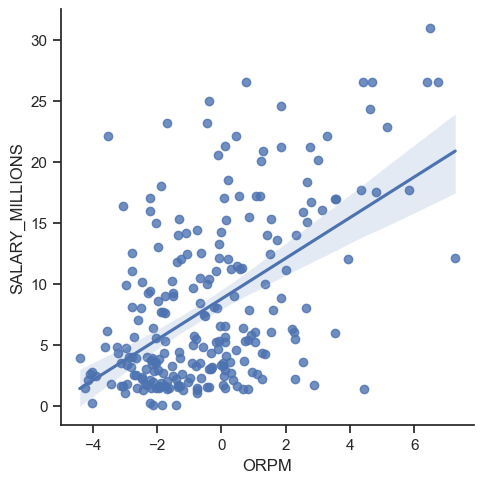}
    \includegraphics[width=8cm]{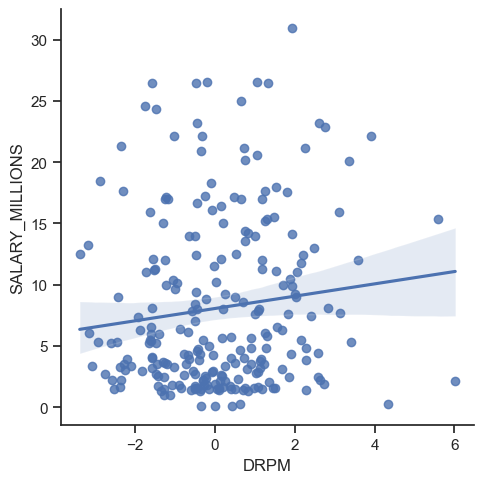}
    \caption{Relations between independent variables and dependent variable, the former consists of MPG/ PTS(for points)/ ORPM/ DRPM, 'MPG' is short for 'minute per game', 'PTS(for points)' is the average PTS(for points) for a player, 'ORPM' represents the offensive real plus-minus, whilst 'DRPM' signifies the defensive real plus-minus. These factors are able to deliver a comprehensive picture to the player's defensive/offensive quality. From the figure, the observation is that 'MPG', 'PTS(for points)' and 'ORPM' are obviously positively correlated with the target variable, which means that outstanding offensive and scoring ability guarantee better salary to some extent. On contrary with that, 'DRPM' shows wealy-connected relation with the target variable, revealing the possible phenomenon that the league has preference for the players with more pronounced offensive quality rather than defensive quality on the average.}
    \label{fig:enter-label}
\end{figure}

In summary, $R^2$ score in joint factor model are much better than independent factor, revealing the indicators are linked intrinsically, comprehensive analysis is preferable compared to figure out the effect of single factor according to this scenario.


\begin{figure}
    \centering
    \includegraphics[width=6cm]{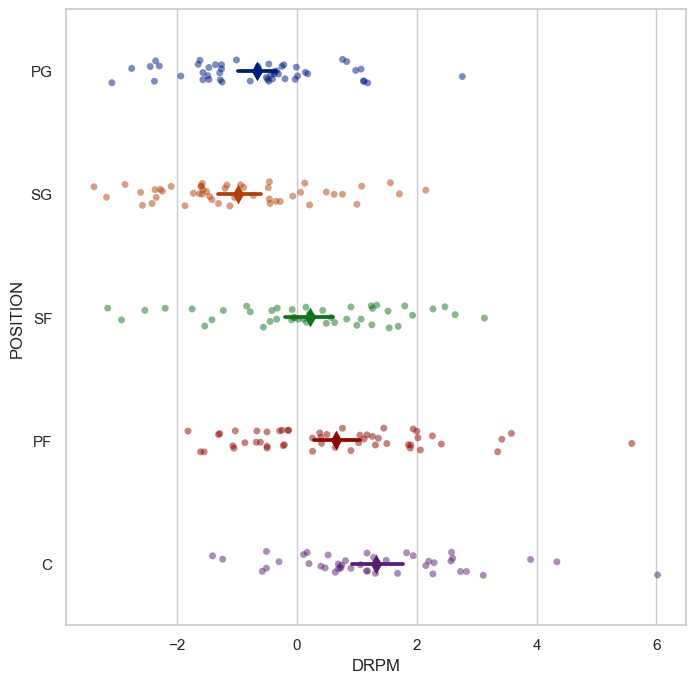}
    \includegraphics[width=9cm]{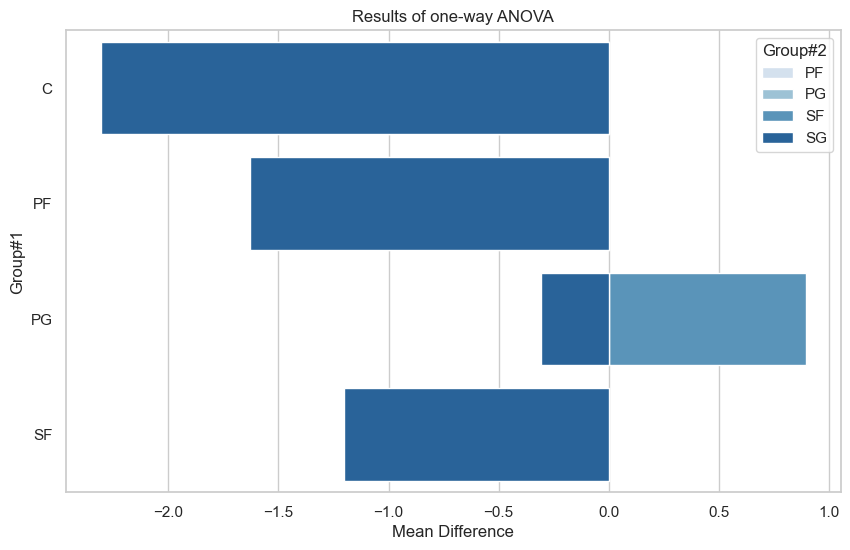}
    \caption{The statistical distribution(left) and analysis(right) of DRPM among varied player roles. Different from the salaries, DRPM, or defensive real plus-minus, has shown great gaps for different player positions. The leading three positions are 'C', 'PF' and 'SF', corresponding to 'Center', 'Pointing Forward' and 'Scoring Forward' respectively. After the iterative updates of p-value subject to TukeyHSD process, among all 10 possible pair-wise comparisons, 7 of them have rejected the original assumption, indicating the discrepancy. In realistic world, these frontcourt roles are responsible for the majority of team defence, hence to have more significant contribution to the DRPM indicator.}
    \label{fig:enter-label}
\end{figure}

\begin{figure}
    \centering
    \vspace{-5pt}
    \includegraphics[width=6cm]{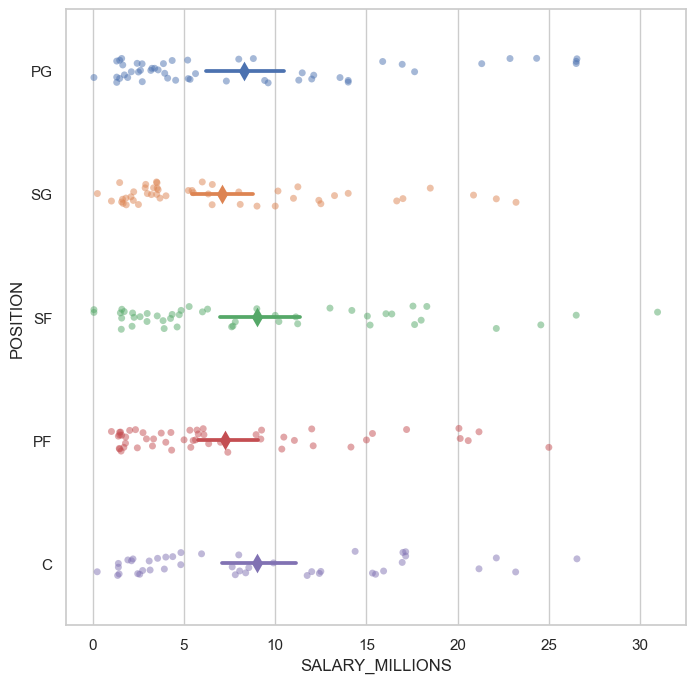}
    \includegraphics[width=9cm]{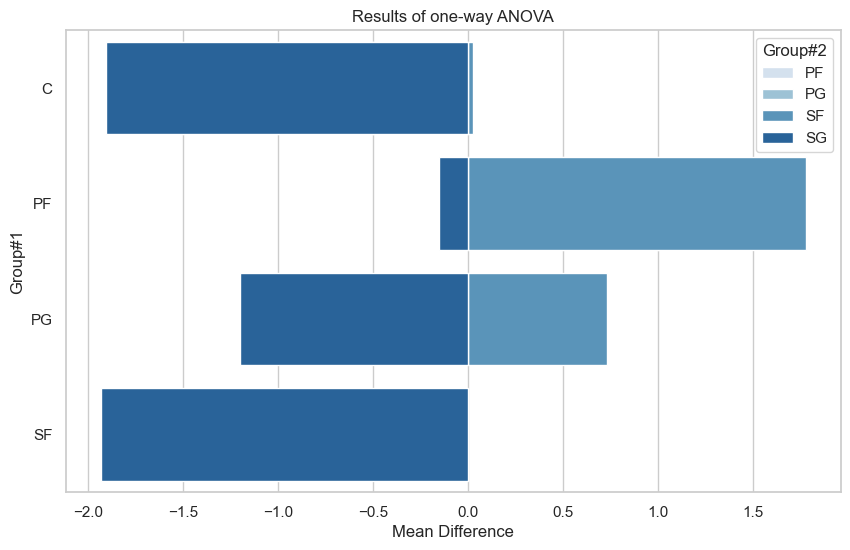}
    \caption{The statistical distribution(left) and analysis(right) of salaries among varied player roles. From the left subplot, there is no significant discrepency between player positions in terms of salaries. To further validate the observation from the statistical perspective, one-way ANOVA(right) is adopted. To be specific, we have conducted pair-wise comparisons followed by the adjustment of p-value. To control the overall familywise error rate(FWER), the lowerbound of FWER is fixed to 0.05, this means that for a set of hypothesis tests, the overall probability of making a mistake is less than 5\%.}
    \label{role-anova}
\end{figure}

\subsubsection{Regression Results}
By incorporating higher-order polynomial features into the \textbf{\textit{LinearRegression}} model, we can capture more complex relationships among the predictor variables. The PolynomialFeatures library is a useful tool for generating these polynomial features. It transforms the original features by considering all possible polynomial combinations up to the specified degree, allowing the model to capture non-linear relationships. In the context of analyzing the high-dimensional NBA dataset, applying PolynomialFeatures with a degree of 2 to the predictor variables (PV(for page views counting), TFC(for twitter favorite counting), TRC(for twitter retweet counting), MPG, PTS(for points), DRPM, ORPM, PN(for position number), AGE) and examining their correlation with the response variable like SM(counting the salary in millions) helps uncover potential quadratic relationships. This enhances the model's ability to explain the variance in salary based on player performance metrics and social power factors. By including polynomial features, the model can account for non-linear effects such as interactions and curvature in the data, providing a more comprehensive understanding of how these predictors influence player salaries.


After incorporating second-order polynomial features into the \textbf{\textit{LinearRegression}} model, we observe a substantial improvement in the model's performance, with an impressive increase in the $R^2$ score from 0.601 to 0.783 . This signifies the importance of considering non-linear relationships when analyzing NBA statistics in relation to salaries. The relationship between player salaries and factors such as historic performance, macro-level circumstances, and league regulations is intricate and cannot be adequately captured by a simple linear model. For instance, it is common for players to receive their highest pay during the middle of their careers, and empirical experiments do not reveal a significant relationship between age and salaries. 

In conclusion, the analysis of NBA salary data requires accounting for non-linear relationships, which can be achieved by incorporating polynomial features into the linear regression model. This approach provides a more accurate representation of the complex dynamics influenced by players' performance history, macro-level factors, and league regulations.

\section{Model Interpretability With Visualization}\label{sec-inter}
Furthermore, the goal of this paper is to construct an interpretable framework to the realistic problems, and make effective predictions. To this end, in this section we are going to excavate the most representative features to indicate player salaries, with the metrics of $R^2$ score quantifying model performance. Meanwhile, plots are presented to better visualize the results, including \textbf{feature effect plot}, \textbf{weight plot}, and \textbf{SHAP}(Shapley Additive exPlanations).

\subsection{Winning Quantities Prediction}
\begin{figure}
    \centering
    \includegraphics[width=14cm]{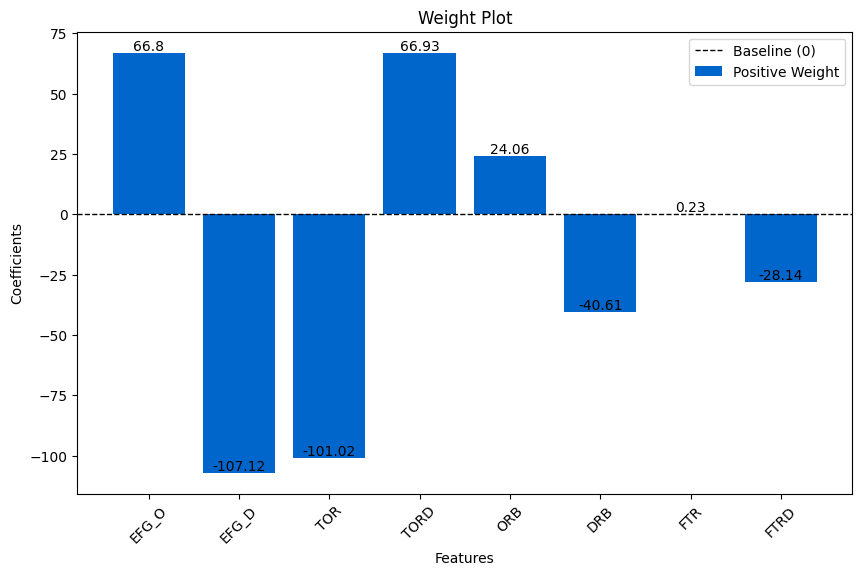}
    \caption{Weight plot of ordinary least squares(OLS). From the figure, we could observe that 'EFG\_O'/'TORD'/'ORB'/'FTR' show positive effects on the target variable, whilst 'EFG\_D'/'TOR'/'DRB'/'FTRD' are negative indicators to the target variable. For the feature importance, 'EFG\_D'/'TOR'/TORD' has the leading three absolute value of weight coefficients.}
    \label{lr-weight}
\end{figure}

To envelope the intrinsic mechanism of linear regression models, we need to visualize the importances of each feature depicted by its weight, and analyze the semantics behind the phenomenon. Weight plot of \textbf{\textit{OLS}} has been placed in the Figure \ref{lr-weight}.

\subsection{Salary v.s. Player Performances}
To improve model interpretability, composition of multivariate variables relations’s formula and feature importance are visualized in the Figure \ref{role-anova}. For factors, points/age/drpm taking the leading roles in the feature importance.

\begin{figure}
    \centering
    \includegraphics[width=15cm]{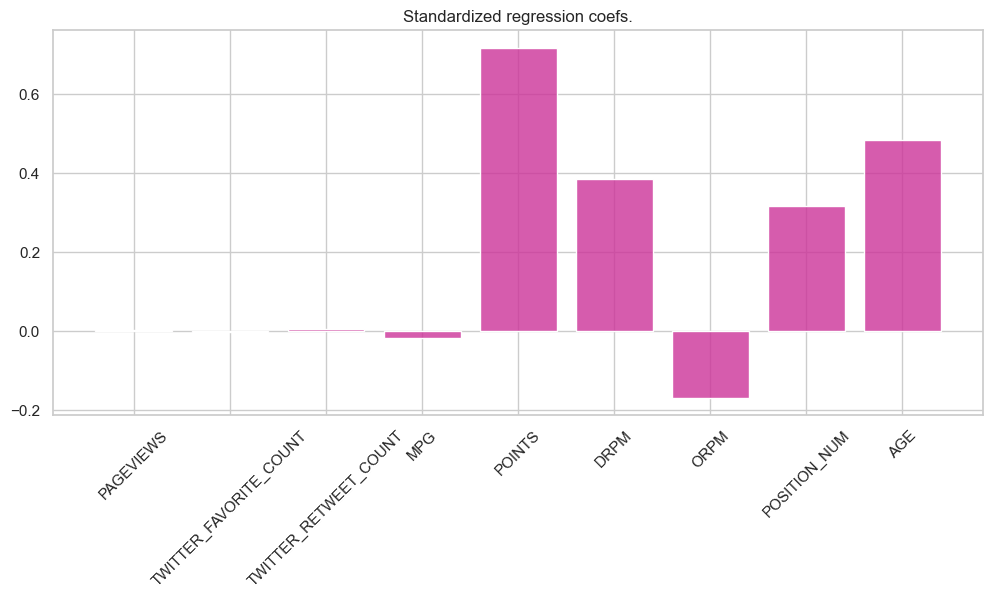}
    \caption{The standardized regression coefficient between predictor variables and a salary variable in regression analysis. From the figure, all predictor variables except TFC(for twitter favorite counting)(-0.0036) and ORPM(-0.0165) show positive coefficient, which indicates a positive relationship.}
    \label{weight plot 2}
\end{figure}

\textbf{Weight Plot}
is proposed in Figure \ref{weight plot 2} to visually present the estimated coefficients of the predictor variables in a linear regression model. Each predictor variable is represented by a point estimate of its coefficient, along with an interval that indicates the uncertainty around the estimate, typically depicted as confidence intervals. The plot provides a visual summary of the magnitude and variability of the estimated effects of each predictor on the response variable.

From the Figure \ref{weight plot 2}, PTS(for points)(0.7180) variable overwhelms the others in view of feature coefficients, followed by AGE(0.4837) and DRPM(0.3847); Social power indicators; Social power factors(PV(the number of page views)(0.0017)/TFC(-0.0036)/TBC(for twitter between counting)(0.0045)) are weak indicators towards salary, the semantics information behind is that well-paied players inclines to have more social explosure, it's positively related, but inversely, when a player has better social power, it's uncertain whether he is a top-player in the league, because salary is a complex of multiple factors, dominated by in-court box statistics rather than others.


\textbf{SHAP(Shapley Additive exPlanations) value}


\subsubsection{Capturing the nonlinearity}
As expection, OLS is extremely insufficient when capturing nonlinearity effects, which gives the birth of generalized additive model(GAM for simplicity), GAM aims to directly learn the non-linear relations by relaxing the constraints that prediction is solely relied on the weighted sum, instead, the outcome can be modeled by arbitrary combination of function mapped from features.

\textbf{Feature Effect}
is also referred to as a partial dependence plot or marginal effect plot, illustrates the relationship between a specific predictor variable and the response variable while holding other variables constant at certain values. Unlike a weight plot, which focuses on individual coefficients, an effect plot enables the visualization of the functional form of the relationship between a predictor and the response.
When creating an effect plot for a linear regression model, the values of the focal predictor variable are typically varied across a range, and the predicted values of the response variable are plotted against these values. This allows for the examination of how changes in the predictor variable are associated with changes in the predicted outcome, providing insights into the nature and direction of the relationship.

\section{Conclusions and Future Work}\label{conclusion}
In this paper, a hybrid interpretable machine learning model was presented to deliver comparative studies in high-dimension sports statistics. The main contribution of this paper would be that we have proposed a brand-new perspective for the data analysis, the feature and the model interpretability should be equally treated. To this end, we have constructed a full pipeline from preprocessing to the result explanation and manual observation. We strongly believe that this framework can not only benefit many downstream industrial applications with the advent of machine learning, but also provide motivation and idea to other researches in term of reliable machine learning, interpretable machine learning, etc.
However, there is still some limitation we haven't overcome. Firstly, the dataset are somehow specific or unique, when carrying the proposed framework, it requires domain knowledge to get all the things work and make it more adaptable. Secondly, in certain cases, the given context is complex enough so we could even explain in the realistic life, so in these circumstances our framework can be partially-interpretable. The concerns and limitations pave the way for the future work.

\section*{Acknowledgments}

\bibliographystyle{unsrt}
\bibliography{references}

\end{document}